\documentclass[letterpaper]{article} 
\usepackage{aaai2027}  
\usepackage[hyphens]{url}  
\usepackage{graphicx} 
\usepackage{natbib}  
\usepackage{caption} 
\usepackage{algorithm}
\usepackage{algorithmic}
\usepackage[table]{xcolor}
\usepackage{multirow}
\usepackage{newfloat}
\usepackage{listings}
\DeclareCaptionStyle{ruled}{labelfont=normalfont,labelsep=colon,strut=off} 
\floatstyle{ruled}
\newfloat{listing}{tb}{lst}{}
\floatname{listing}{Listing}

\usepackage{booktabs}
\usepackage{tabularx}
\usepackage{array}
\title{RegionDet: A Benchmark for Region Detection Beyond Object Instances}
\author{
    Liang Wan\textsuperscript{\rm 1,2}, 
    Yuhan Wang\textsuperscript{\rm 1},
    Yupeng Zhang\textsuperscript{\rm 1,2}\corresponding,
    Zhen Xu\textsuperscript{\rm 1},
    Han Wang\textsuperscript{\rm 1},
    Fangjie Fu\textsuperscript{\rm 1}, 
    Sirui Zhu\textsuperscript{\rm 1}
}
\affiliations{
	\textsuperscript{\rm 1}College of Intelligence and Computing, Tianjin University.\\
    \textsuperscript{\rm 2}Key Research Center for Surface Monitoring and Analysis of Relics, State Administration of Cultural Heritage.\\
}

\begin{document}

\maketitle

\begin{abstract}
Object detection is a fundamental task in computer vision and has achieved remarkable progress on standard benchmarks by localizing discrete and well-bounded object instances. 
However, many visual targets in real-world scenarios are not individual objects, but regions defined by visual states, scene context, object relations, and human activities, such as construction areas, damaged road regions, queues, group conversations, and vendor regions. 
Existing detection benchmarks are mainly built around object instances, providing limited support for systematically evaluating such region targets.
To address this gap, we introduce \textbf{Region Detection}, a task that extends conventional object detection beyond object instances, and construct \textbf{RegionDet}, a benchmark for region target localization. 
RegionDet contains eight region categories, including Construction, Crossing, Damage, Queuing, Talking, Vendor, Waiting, and Walking, with COCO-style bounding-box annotations and evaluation protocols. 
We systematically evaluate representative closed-set and zero-shot/open-vocabulary detectors on RegionDet. 
Results show that closed-set detectors can partially learn region-level patterns under supervision, while zero-shot/open-vocabulary detectors struggle severely, revealing the strong object-centric bias of current vision-language detectors. 
Further analyses highlight key challenges in Region Detection, including weak boundary cues, strong context dependency, and insufficient relation-level region understanding. 
\textit{The RegionDet will be released.}
\end{abstract}



\section{Introduction}
Object detection is a fundamental task in computer vision, aiming to localize and recognize semantic visual targets in images. 
With the rapid development of modern detectors, remarkable progress has been achieved on standard benchmarks such as MS COCO~\cite{lin2014microsoft} and LVIS~\cite{gupta2019lvis}. 
However, most existing detection paradigms are built upon an object-centric assumption, where targets are typically discrete and countable entities with relatively clear physical boundaries, such as people, vehicles, animals, and everyday objects. 
Although this object-centric setting has greatly advanced generic object detection, it does not fully cover the broader visual localization needs of real-world vision systems.

\begin{figure}[t!]
	\centering
	\includegraphics[width=\linewidth]{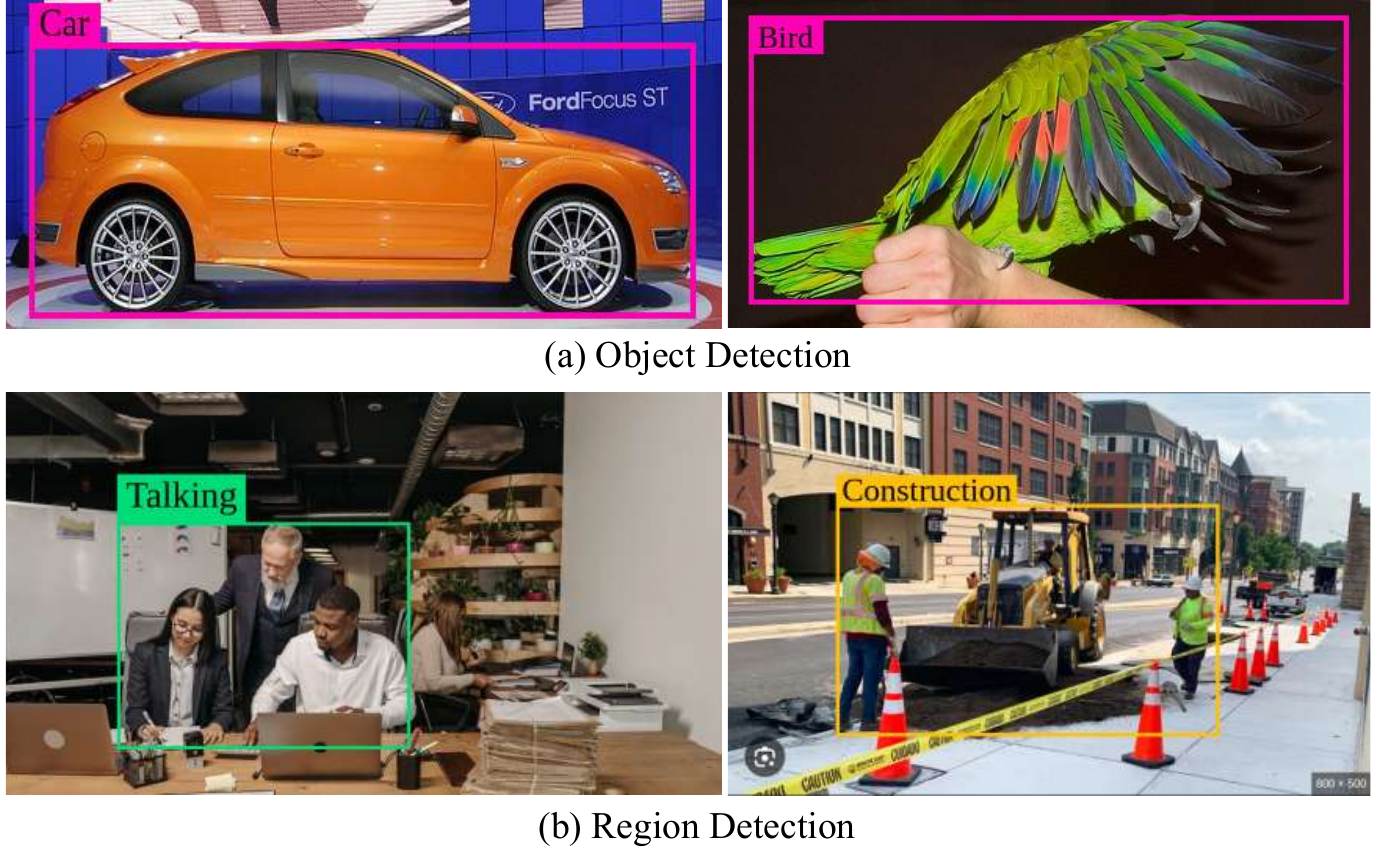}	
    \vspace{-20pt}
    \caption{Comparison between conventional object detection and Region Detection. (a) Object detection focuses on discrete and well-bounded object instances. (b) Region Detection localizes region targets whose spatial extents are defined by visual states, object relations, scene functions, and contextual cues beyond single-object boundaries.}
	\label{fig:comp}
	\vspace{-15pt}
\end{figure}

In many real-world applications, visual systems need to localize not only individual object instances, but also visual regions with semantic, state, functional, or event-level meanings. 
For example, in urban streets, public spaces, and surveillance scenarios, targets of interest may include construction areas, damaged road regions, queues, group conversations, and vendor regions. 
These targets do not necessarily correspond to single standalone objects; instead, they are often defined by visual states, spatial relations among multiple objects, local scene functions, and surrounding context, and usually lack stable physical boundaries. 
For instance, a construction area is not determined by the boundary of a worker, vehicle, or machine, but by workers, equipment, material piles, road context, and ongoing activities. 
Similarly, a group conversation is not a single person box, but is determined by the proximity, pose orientation, and interaction among multiple participants. 
Fig.~\ref{fig:comp} illustrates the difference between conventional object detection and Region Detection: the former mainly focuses on well-bounded object instances, while the latter localizes region targets defined by visual states, object relations, scene functions, and contextual cues.

These region targets differ fundamentally from conventional object instances in three aspects:
\textbf{(1) Weak boundary cues.} Their spatial extents are often not directly defined by clear object contours, making localization difficult with standard boundary cues;
\textbf{(2) Diverse forms.} They may appear as local abnormal regions or activity areas formed by multiple people or objects;
\textbf{(3) Strong context dependency.} Many region targets rely heavily on scene context and object relations, and thus cannot be reliably recognized from local appearance alone.
This raises a key question: \textbf{\textit{when detection targets are no longer well-bounded, discrete, and countable object instances, but regions defined by visual states, scene context, and object interactions, can existing object detectors still localize and recognize them reliably?}}

To answer this question, we introduce and study \textbf{Region Detection}, a task that aims to localize region targets with semantic, state, functional, or event-level meanings beyond conventional object instances. 
Unlike conventional object detection, which mainly focuses on discrete and countable entities, Region Detection extends detection targets to visual regions defined by behavioral states, group relations, scene functions, and local abnormalities. 
This setting preserves the basic localization-and-classification formulation of detection, while better matching the region localization needs in real-world applications such as traffic monitoring, urban management, public safety, and scene understanding.

Based on this setting, we construct \textbf{RegionDet}, a benchmark dataset for Region Detection. 
RegionDet contains eight region categories, including Construction, Crossing, Damage, Queuing, Talking, Vendor, Waiting, and Walking, covering typical scenarios such as construction activities, pedestrian behaviors, group interactions, vendor scenes, waiting states, and local damage. 
To ensure a unified task definition and reproducible evaluation, we annotate RegionDet with MS COCO-compatible bounding boxes and category labels, and establish standard data splits and evaluation protocols. 
This annotation format allows RegionDet to be directly adapted to existing object detection frameworks, while providing a basis for systematically analyzing the capability boundaries of current detectors on region targets beyond conventional object instances.

Furthermore, we systematically evaluate representative closed-set and zero-shot/open-vocabulary detectors across diverse detection paradigms, including two-stage, one-stage, point/center-based, diffusion-based, query-based Transformer, and vision-language detection. 
Experimental results show that RegionDet poses new challenges to existing detectors. 
Closed-set detectors can partially learn region-level patterns under supervised training, but their performance varies notably across paradigms. 
In contrast, zero-shot/open-vocabulary detectors are clearly limited, suggesting that current vision-language detection models remain strongly object-centric and struggle to generalize to region targets defined by states, relations, and context.

Our main contributions are summarized as follows:

\begin{itemize}
    \item We introduce \textbf{Region Detection}, a detection task that extends targets from well-bounded, discrete, and countable object instances to region targets defined by visual states, scene context, and object interactions.
    \item We construct \textbf{RegionDet}, a benchmark dataset containing eight region categories: Construction, Crossing, Damage, Queuing, Talking, Vendor, Waiting, and Walking, with MS COCO-compatible bounding-box annotations and evaluation protocols.
    \item We systematically evaluate representative closed-set and zero-shot/open-vocabulary detectors, covering major paradigms including two-stage, one-stage, point/center-based, diffusion-based, query-based Transformer, and vision-language detection.
    \item We reveal key limitations of existing detectors on Region Detection, including weak boundary cues, strong context dependency, and insufficient relation-level region understanding, and further show that current zero-shot/open-vocabulary detectors remain strongly object-centric.
\end{itemize}

\section{Related Work}

\textbf{Object detection benchmarks.}
Large-scale benchmarks have played a central role in the development of object detection. 
Early datasets such as PASCAL VOC~\cite{everingham2010pascal} established the standard formulation of localizing object instances with bounding boxes and category labels. 
MS COCO~\cite{lin2014microsoft} further advanced detection research by introducing more diverse scenes, object scales, and instance annotations. 
Later benchmarks and benchmark suites, including Open Images~\cite{kuznetsova2020open}, Objects365~\cite{shao2019objects365}, LVIS~\cite{gupta2019lvis}, ODinW~\cite{li2022elevater}, and MSOSB~\cite{zhang2024rethinking}, expanded the category vocabulary, data scale, long-tailed distribution, or domain coverage of object detection.
Despite these advances, existing object detection benchmarks are still largely object-centric. 
Their annotated targets are typically discrete, countable, and well-bounded object instances, such as persons, vehicles, animals, and everyday objects. 
This formulation has greatly advanced generic object detection, but provides limited support for studying targets whose spatial extents are defined by visual states, scene context, group relations, or activities. 
In contrast, RegionDet focuses on region targets beyond conventional object instances, such as construction areas, damaged regions, queues, group conversations, and vendor regions. 
These targets are not necessarily determined by the physical contour of a single object, and thus require detectors to reason about contextual and relational cues beyond standard object boundaries.

\textbf{Region-level visual understanding.}
Beyond object detection, several vision tasks aim to understand visual regions at a finer or broader semantic level. 
Semantic segmentation datasets, such as PASCAL Context~\cite{mottaghi2014role}, ADE20K~\cite{zhou2017scene}, and Cityscapes~\cite{cordts2016cityscapes}, assign semantic labels to pixels and have enabled dense scene parsing. 
Stuff and panoptic segmentation benchmarks, including COCO-Stuff~\cite{caesar2018coco} and COCO-Panoptic~\cite{kirillov2019panoptic}, further extend scene understanding from object instances to amorphous regions such as road, sky, grass, and sidewalk. 
These benchmarks highlight the importance of modeling regions beyond object instances.
However, Region Detection differs from segmentation-based scene understanding in both target definition and annotation form. 
Segmentation benchmarks typically focus on pixel-level labeling of continuous object or stuff categories, while RegionDet adopts bounding boxes to localize semantic, functional, state, or activity regions. 
Many region targets in RegionDet, such as \textit{Queuing}, \textit{Talking}, and \textit{Vendor}, may involve multiple people, objects, and contextual elements, and do not necessarily correspond to a single continuous mask region. 
Therefore, RegionDet preserves the efficient box-level detection formulation while extending the target space toward context-dependent and relation-defined visual regions.
RegionDet is also related to action and interaction understanding. 
Datasets such as AVA~\cite{gu2018ava}, V-COCO~\cite{gupta2015visual}, and HICO-DET~\cite{chao2018learning} study human actions or human-object interactions. 
However, these tasks usually focus on recognizing individual actions, actor boxes, or human-object-relation triplets. 
In contrast, RegionDet aims to localize the spatial extent of an activity, functional, or interaction region, which may include multiple participants, objects, and surrounding context. 
Thus, RegionDet provides a complementary benchmark for evaluating whether detection models can move from object-level localization to region-level visual understanding.

\textbf{Open-vocabulary and vision-language detection.}
Recent open-vocabulary and vision-language detectors aim to recognize novel categories specified by text descriptions. 
Representative methods, such as GLIP~\cite{li2022grounded}, Grounding DINO~\cite{liu2024grounding}, OWL-ViT~\cite{minderer2022simple}, YOLO-World~\cite{cheng2024yolo}, YOLOE~\cite{wang2025yoloe}, and LLMDet~\cite{fu2025llmdet}, leverage large-scale image-text pretraining or grounding data to align visual regions with language queries. 
These methods have shown promising generalization ability on conventional object detection and phrase grounding tasks.
Nevertheless, most existing open-vocabulary detectors are still primarily trained and evaluated on object-like targets. 
Their language queries often correspond to nouns or noun phrases associated with individual object instances, such as `person', `car', or `traffic light'. 
It remains unclear whether such models can directly generalize to region targets defined by states, activities, group relations, and scene context. 
RegionDet provides a benchmark to evaluate this capability. 
Our experiments show that current zero-shot/open-vocabulary detectors struggle severely on RegionDet, suggesting that existing vision-language detection models remain strongly object-centric and are not yet sufficient for localizing region targets beyond object instances.

\section{Region Detection Task and RegionDet Benchmark}
\label{sec:regiondet}

\subsection{Task Definition}

We define \textbf{Region Detection} as a detection task that aims to localize visual regions with semantic, state, functional, or event-level meanings beyond conventional object instances. 
Given an input image $I$, the goal is to predict a set of region targets:
\begin{equation}
    \hat{\mathcal{Y}} = \{(\hat{b}_i, \hat{c}_i, \hat{s}_i)\}_{i=1}^{N},
\end{equation}
where $\hat{b}_i = (\hat{x}_i, \hat{y}_i, \hat{w}_i, \hat{h}_i)$ denotes the predicted bounding box, $\hat{c}_i \in \mathcal{C}$ is the predicted region category, $\hat{s}_i$ is the confidence score, and $N$ is the number of predicted region targets. 
The category set $\mathcal{C}$ consists of region categories rather than conventional object categories.

For each image, the ground-truth annotations are represented as:
\begin{equation}
    \mathcal{Y} = \{(b_j, c_j)\}_{j=1}^{M},
\end{equation}
where $b_j$ and $c_j$ denote the bounding box and category label of the $j$-th annotated region, respectively. 
Similar to standard object detection, Region Detection adopts bounding boxes and category labels as the basic annotation format, making it compatible with existing detection frameworks and evaluation metrics. 
However, unlike conventional object detection, region targets are not necessarily defined by the physical contour of a single object. 
Instead, their spatial extents are often determined by visual states, object interactions, group relations, scene functions, and surrounding context.

\begin{figure}[t]
    \centering
    \includegraphics[width=\linewidth]{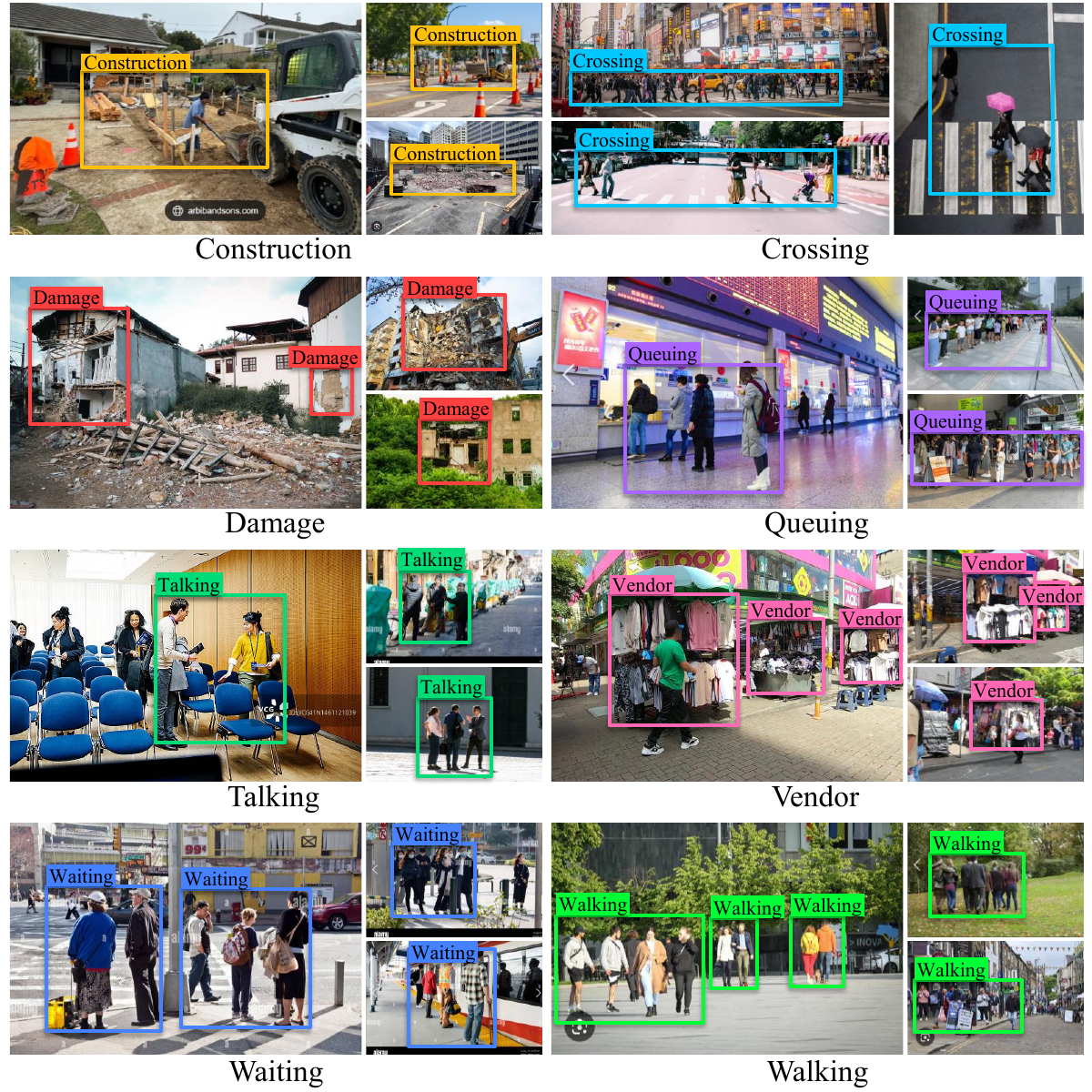}
    \vspace{-20pt}
    \caption{Representative examples of RegionDet. Region targets may be defined by visual states, object relations, scene functions, and contextual cues rather than the physical contour of a single object.}  
    \label{fig:exam}
    \vspace{-15pt}
\end{figure}

This formulation preserves the standard localization-and-classification paradigm of object detection, while extending the target space from well-bounded object instances to semantically meaningful region targets. 
Table~\ref{tab:benchmark_comparison} summarizes the differences between RegionDet and related benchmarks. 
Fig.~\ref{fig:exam} shows representative examples in RegionDet.

\begin{table*}[t!]
\centering
\scriptsize
\setlength{\tabcolsep}{2.5pt}
\caption{Comparison between RegionDet and related benchmarks.}
\vspace{-10pt}
\begin{tabular}{l|c|c|c|c}
\toprule
Benchmark & Annotation & Target Type & Context/Relation & Box-level Region Detection \\
\midrule
COCO/LVIS & Box/Mask & Object instances & Limited & No \\
ADE20K/COCO-Stuff & Mask & Stuff/scene regions & Limited & No \\
AVA/HICO-DET & Box/Triplet & Actions/interactions & Yes & Partial \\
Referring Segmentation & Mask & Language-referred regions/objects & Yes & No \\
RegionDet & Box & Region targets beyond objects & Yes & Yes \\
\bottomrule
\end{tabular}
\label{tab:benchmark_comparison}
\vspace{-10pt}
\end{table*}

\subsection{Region Categories}

RegionDet contains eight region categories: \textit{Construction}, \textit{Crossing}, \textit{Damage}, \textit{Queuing}, \textit{Talking}, \textit{Vendor}, \textit{Waiting}, and \textit{Walking}. 
These categories cover common region-level targets in urban streets, public spaces, and surveillance scenarios, including activity regions, abnormal regions, group interaction regions, and functional regions.

Unlike conventional object categories such as person, car, or bicycle, these region categories are often defined by spatial relations and contextual cues. 
For example, a \textit{Talking} region is determined by interactions among multiple participants rather than the bounding box of a single person, while a \textit{Construction} region is jointly formed by workers, equipment, materials, road context, and ongoing construction activities. 
Table~\ref{tab:category_definition} provides the detailed definition of each category.



\begin{table}[t]
\centering
\scriptsize
\setlength{\tabcolsep}{4pt}
\renewcommand{\arraystretch}{1.05}
\caption{Category definitions of RegionDet.}
\vspace{-10pt}
{
\renewcommand{\tabularxcolumn}[1]{m{#1}}
\begin{tabularx}{\linewidth}{>{\raggedright\arraybackslash}m{1.75cm}|>{\raggedright\arraybackslash}X}
\toprule
Category & Definition \\
\midrule
Construction 
& Regions where construction-related activities are taking place, typically involving workers, equipment, materials, and road context. \\
\midrule
Crossing 
& Regions where pedestrians are crossing roads or moving across traffic-related spaces, usually defined by pedestrian behavior and road context. \\
\midrule
Damage 
& Local abnormal or damaged regions, such as road damage, surface breakage, or visually degraded areas. \\
\midrule
Queuing 
& Regions where multiple people form a queue or line, determined by their spatial arrangement and waiting behavior. \\
\midrule
Talking 
& Regions where multiple people are engaged in conversation or social interaction, defined by their proximity, pose orientation, and interaction patterns. \\
\midrule
Vendor 
& Functional regions related to street vending or trading, typically involving vendors, goods, customers, and the local trading context. \\
\midrule
Waiting 
& Regions where people are waiting or staying in place, usually inferred from posture, spatial arrangement, and scene context. \\
\midrule
Walking 
& Regions where pedestrians are walking or moving, defined by human motion states and the surrounding scene context. \\
\bottomrule
\end{tabularx}
}
\label{tab:category_definition}
\vspace{-15pt}
\end{table}

\subsection{Data Collection and Annotation}

To construct RegionDet, we collected images from public web sources covering diverse real-world scenarios, including urban streets, public spaces, traffic scenes, and surveillance-like environments. 
Ten annotators were responsible for image collection, filtering, and initial bounding-box annotation. 
During data collection, images were selected to ensure that the target regions were visually recognizable and that the scenes covered diverse viewpoints, scales, densities, and contextual layouts. 
Low-quality, duplicated, irrelevant, or highly ambiguous images were removed.

Each target region was annotated with a bounding box and one of the eight predefined region categories. 
Since region targets may not correspond to the contour of a single object, annotators were instructed to annotate the spatial extent that best covers the intended semantic, functional, or activity region. 
For example, a \textit{Talking} region should include the interacting participants that jointly form the conversation rather than only one person, while a \textit{Construction} region should cover the local construction activity area, including relevant workers, equipment, materials, and contextual elements when necessary. 
For \textit{Damage}, annotators focused on the visible abnormal or damaged area; for \textit{Vendor}, the annotated region covered the functional trading area formed by the vendor, goods, customers, and nearby context.

To improve annotation quality, five additional annotators conducted two rounds of verification. 
The first round checked whether category labels were correct and whether the boxes covered the intended region targets. 
The second round further inspected ambiguous cases, removed low-quality annotations, and refined boxes with inaccurate spatial extents. 
Disagreements and uncertain cases were discussed and resolved according to the annotation guidelines. 
This two-round verification process improves the consistency and reliability of RegionDet annotations.




\begin{table}[t!]
\centering
\scriptsize
\setlength{\tabcolsep}{1.2pt}
\caption{Category-wise statistics of RegionDet.}
\vspace{-10pt}
\begin{tabular}{l|cc|cc|cc}
\toprule
Category 
& Images & Regions 
& Train Images & Train Regions 
& Test Images & Test Regions \\
\midrule
Construction & 1999 & 2031  & 1000 & 1016 & 999 & 1015 \\
Crossing     & 401  & 415   & 201  & 208  & 200 & 207  \\
Damage       & 1999 & 4739  & 1000 & 2437 & 999 & 2302 \\
Queuing      & 409  & 426   & 204  & 212  & 205 & 214  \\
Talking      & 406  & 574   & 203  & 257  & 203 & 317  \\
Vendor       & 1995 & 3301  & 997  & 1673 & 998 & 1628 \\
Waiting      & 400  & 571   & 200  & 320  & 200 & 251  \\
Walking      & 401  & 531   & 200  & 248  & 201 & 283  \\
\midrule
Total        & 8010 & 12588 & 4005 & 6371 & 4005 & 6217 \\
\bottomrule
\end{tabular}
\label{tab:category_statistics}
\vspace{-15pt}
\end{table}

\begin{table*}[t!]
\centering
\scriptsize
\setlength{\tabcolsep}{2.5pt}
\caption{Results of Closed-set Detection Methods on RegionDet (\%).} \vspace{-10pt}
\begin{tabular}{l|l|c|cccccccc|ccc|c}
\toprule
Type &Methods & Backbone/Encoder  & Construction & Crossing & Damage & Queuing & Talking & Vendor & Waiting & Walking & AP & AP$_{50}$ & AP$_{75}$ & Params. (M) \\
\hline
\multirow{3}{*}{\shortstack[l]{Dense\\one-stage}} 
&ATSS & ResNet-50 &28.5 &18.5 &6.0 &13.9 &19.2 &19.4 &9.0 &17.7 &16.5 &35.7 &13.0 &32.1\\
&YOLOv13-L & YOLOv13 &31.8&54.0&6.61&30.7&33.7&24.8&15.6&40.2&29.7&48.5&29.8 &27.6\\
&YOLO26-L & YOLO26 &33.3&53.2&8.2&28.3&33.7&25.1&18.0&44.6&30.5&49.2&31.4 &24.8\\
\hline
Two-stage 
&Faster R-CNN  & ResNet-50 &25.5 &13.1 &5.7 &16.5 &17.7 &18.2 &8.8 &20.0 &15.7 &38.1 &10.3 &41.4\\
\hline
\multirow{2}{*}{\shortstack[l]{Point/center\\anchor-free}}
&RepPoints & ResNet-50 &28.8 &17.6 &6.8 &15.9 &19.4 &19.9 &9.6 &17.5 &16.9 &36.7 &13.3 &36.8\\
&CenterNet++  & ResNet-50 &34.6 &35.8 &6.8 &22.5 &21.8 &17.8 &13.4 &26.6 &22.4 &40.8 &22.3 &40.8\\
\hline
\multirow{2}{*}{\shortstack[l]{Diffusion\\based}} 
&DiffusionDet & ResNet-50 & 29.7 & 40.8 & 6.1 & 23.0 & 21.5 & 18.6 & 17.4 & 30.6 & 23.5 & 44.5 & 22.3 & 110.6 \\
&DiffuDETR  &ResNet-50 &32.1 &37.5 &6.5 &23.2 &20.1 &18.6 &16.8 &27.5 &22.8 &39.9 &22.9 &59.1\\  
\hline
\multirow{6}{*}{\shortstack[l]{Query-based\\Transformer}} 
&DEIMv2-L &DINOv3-S16 &45.5&67.7&15.5&31.4&37.0&35.0&25.3&38.8&37.0&57.1&38.7 &32.0\\
&RT-DETRv4-L &DINOv3-B16 &38.6&62.7&11.3&32.2&35.5&29.0&22.5&32.8&33.1&52.7&34.7 &31.0 \\
&RF-DETR-L &DINOv2-S16 &37.5 &60.0 &13.3 &32.8 & 43.7 &34.5 &24.5 & 37.8 &37.6 &56.1 &37.4 &33.6\\
\cline{2-15}
&DINO & ResNet-50 &29.7 &36.6 &5.7 &17.8 &15.7 &15.0 &10.7 &23.2 &19.3 &34.0 &19.1 &47.6\\ 
&Co-DETR & ResNet-50 &35.4 &46.8 &7.8 &25.5 &25.0 &22.3 &17.3 &31.9 &26.5 &46.2 &26.2 &64.5\\
&MI-DETR  &ResNet-50 &33.6 &47.0 &6.9 &23.9 &22.2 &20.1 &15.2 &31.9 &25.1 &44.4 &25.0 &77.3\\
\bottomrule
\end{tabular}\vspace{-15pt}
\label{tab:close}
\end{table*}

\subsection{Dataset Statistics}
RegionDet contains 8,010 images and 12,588 annotated region targets in total. 
The dataset is split into a training set and a test set, with 4,005 images and 6,371 regions for training, and 4,005 images and 6,217 regions for testing. 
Table~\ref{tab:category_statistics} reports the category-wise distribution and overall statistics of RegionDet.
RegionDet covers eight region categories with diverse semantic properties and spatial characteristics. 
For categories with larger appearance variations and scene diversity, such as \textit{Construction}, \textit{Damage}, and \textit{Vendor}, we collected relatively more samples. 
Interaction- and behavior-related categories such as \textit{Queuing}, \textit{Talking}, \textit{Waiting}, and \textit{Walking} are included to evaluate relation- and state-aware region understanding. 
The training and test sets are approximately balanced within each category, enabling consistent evaluation across region targets.

\subsection{Evaluation Protocol}
We evaluate detectors on the RegionDet test set using standard COCO-style detection metrics. 
The primary metric is mean Average Precision (AP), computed over IoU thresholds from 0.50 to 0.95 with a step size of 0.05 and averaged over all categories. 
We also report AP$_{50}$ and AP$_{75}$ to evaluate localization performance under different overlap requirements.

For closed-set detection, all models are trained on the RegionDet training set and evaluated on the test set under the same eight-category label space. 
For zero-shot/open-vocabulary detection, models are evaluated without using any RegionDet training data. 
We use category names as text queries and evaluate the predicted boxes under the same COCO-style protocol. 
The text queries are kept fixed across methods for fair comparison. 
This unified setup enables direct comparison between supervised detectors and zero-shot/open-vocabulary detectors on region targets.

\section{Empirical Results and Findings}
\label{sec:experiments}

\subsection{Experimental Setup}

\textbf{Evaluation settings.}
We evaluate existing detectors on RegionDet under two settings: closed-set detection and zero-shot/open-vocabulary detection. 
In the closed-set setting, models are trained on the RegionDet training set and evaluated on the test set with the same eight region categories, assessing whether existing detection architectures can learn region targets under full supervision. 
In the zero-shot/open-vocabulary setting, models are evaluated without using RegionDet training data. 
We use the RegionDet category names as fixed text queries across methods to examine whether vision-language detectors trained on large-scale image-text or grounding data can generalize to region targets beyond conventional object instances.

\textbf{Evaluation metrics.}
Following the evaluation protocol in \textit{Section Evaluation Protocol}, we report AP, AP$_{50}$, AP$_{75}$, and category-wise AP for all eight region categories. 
Model parameters are reported in millions when available.

\textbf{Evaluated methods.}
For closed-set detection, we evaluate detectors from diverse paradigms, as summarized in Table~\ref{tab:close}. 
These include dense one-stage detectors (ATSS~\cite{zhang2020bridging}, YOLOv13-L~\cite{lei2025yolov13}, and YOLO26-L~\cite{sapkota2025yolo26}), a two-stage detector (Faster R-CNN~\cite{ren2016faster}), point/center-based anchor-free detectors (RepPoints~\cite{yang2019reppoints} and CenterNet++~\cite{duan2023centernet++}), diffusion-based detectors (DiffusionDet~\cite{chen2023diffusiondet} and DiffuDETR~\cite{nawar2026diffudetr}), and query-based Transformer detectors (DINO~\cite{zhang2022dino}, Co-DETR~\cite{zong2023detrs}, MI-DETR~\cite{nan2025mi}, RF-DETR-L~\cite{robinson2025rf}, RT-DETRv4-L~\cite{liao2025rt}, and DEIMv2-L~\cite{huang2025real}). 
For zero-shot/open-vocabulary detection, we evaluate text-conditioned vision-language detectors, including MMGroundingDINO~\cite{liu2024grounding}, YOLO-World~\cite{cheng2024yolo}, YOLOE~\cite{wang2025yoloe}, and LLMDet~\cite{fu2025llmdet}, as shown in Table~\ref{tab:zero-shot}. 
We group methods by their dominant detection paradigm, although some architectural designs may overlap across categories.

\begin{table*}[t!]
\centering
\scriptsize
\setlength{\tabcolsep}{2.8pt}
\caption{Results of Zero-shot/Open-vocabulary Detectors on RegionDet (\%).}
\vspace{-10pt}
\begin{tabular}{l|c|cccccccc|ccc|c}
\toprule
Methods & Backbone/Encoder  
& Construction & Crossing & Damage & Queuing & Talking & Vendor & Waiting & Walking 
& AP & AP$_{50}$ & AP$_{75}$ & Params. (M) \\
\hline
\multirow{3}{*}{MMGroundingDino} 
& Swin-T  & 3.4 & 0.2 & 1.2 & 0.0 & 0.3 & 0.6 & 0.3 & 0.2 & 0.8 & 1.9 & 0.6 &173 \\
& Swin-B  & 1.1 & 0.1 & 0.6 & 0.0 & 0.3 & 0.4 & 0.2 & 0.2 & 0.3 & 1.1 & 0.2 &233 \\
& Swin-L & 1.4 & 0.3 & 0.1 & 0.0 & 0.5 & 3.6 & 0.3 & 0.3 & 0.8 & 2.1 & 0.6 &343 \\
\hline
\multirow{3}{*}{YOLO World} 
& YOLOv8-S &2.3  &0.1  &0.0  &0.0  &0.4  &0.1  &0.1  &0.0  & 0.4 & 1.1 & 0.2 &76 \\
& YOLOv8-M &3.6  &0.5  &0.0  &0.1  &0.5  &0.1  &0.8  &0.1  & 0.7 & 1.9 & 0.4 &92 \\
& YOLOv8-L &3.5  & 0.0 &0.0  &0.0  &1.0  &0.1  &0.2  & 0.0 & 0.6 & 1.8 & 0.3 &110 \\
\hline
\multirow{6}{*}{YOLOE}
& YOLOv8-S  &4.2  &0.1  &0.3  &0.7  &0.2  &0.3  &0.3  &0.2  & 0.7 & 2.0 & 0.4 &78 \\
& YOLOv8-M  &4.6  &0.4  &0.3  &0.5  &0.2  &0.4  &0.2  &0.3  & 0.9 & 2.5 & 0.5 &95 \\
& YOLOv8-L  &5.1  &0.4  &0.4  &0.5  &0.3  &0.5  &0.4  &0.1  & 1.0 & 2.8 & 0.5 &115 \\
\cline{2-14}
& YOLOv11-S &4.0  &0.4  &0.2  &0.2  &0.3  &0.3  &0.2  &0.3  & 0.7 & 2.2 & 0.4 &76 \\
& YOLOv11-M &4.9  &0.2  &0.2  &0.5  &0.1  &0.3  &0.4  &0.2  & 0.8 & 2.4 & 0.4 &91 \\
& YOLOv11-L &4.5  &0.1  &0.9  &0.6  &0.1  &0.4  &0.6  &0.0  & 0.9 & 2.5 & 0.5 &96 \\
\hline
\multirow{3}{*}{LLMDet} 
& Swin-T & 0.7 & 0.1 & 0.8 & 0.0 & 0.4 & 0.2 & 0.1 & 0.1 & 0.3 & 0.9 & 0.2 &744 \\
& Swin-B & 1.0 & 0.2 & 0.8 & 0.0 & 0.3 & 0.7 & 0.1 & 0.1 & 0.4 & 1.1 & 0.2 &804 \\
& Swin-L & 0.8 & 0.2 & 0.4 & 0.0 & 0.4 & 0.2 & 0.1 & 0.1 & 0.3 & 0.9 & 0.2 &879 \\
\bottomrule
\end{tabular}
\vspace{-15pt}
\label{tab:zero-shot}
\end{table*}

\subsection{Closed-set Detection Results}

Table~\ref{tab:close} reports the results of closed-set detectors on RegionDet. 
Overall, existing detectors can learn region-level patterns under full supervision, but RegionDet remains challenging even for recent strong detectors. 
Among all evaluated methods, RF-DETR-L achieves the best overall AP of 37.6, while DEIMv2-L obtains a comparable AP of 37.0 and the best AP$_{50}$ and AP$_{75}$ scores. 
These results suggest that query-based Transformer detectors with strong visual encoders are more effective for Region Detection.

\textbf{Performance varies notably across detection paradigms.} 
Dense one-stage and two-stage detectors show relatively limited performance, with ATSS and Faster R-CNN achieving 16.5 AP and 15.7 AP, respectively. 
Point/center-based anchor-free and diffusion-based detectors perform better, with CenterNet++ and DiffusionDet reaching 22.4 AP and 23.5 AP. 
Query-based Transformer detectors generally achieve stronger results. 
For example, Co-DETR improves over DINO by 7.2 AP, while RF-DETR-L and DEIMv2-L further improve the performance to around 37 AP. 
This suggests that global query interactions, strong visual representations, and flexible set prediction are beneficial for region-level localization.

\textbf{Category-wise results reveal large difficulty differences among region targets.} 
\textit{Crossing} is relatively easier for strong detectors, with DEIMv2-L reaching 67.7 AP and RF-DETR-L reaching 60.0 AP, likely because crossing regions often contain salient pedestrian-road context and recognizable spatial layouts. 
In contrast, \textit{Damage} is consistently the most difficult category. 
Even the best-performing method achieves only 15.5 AP on this category, indicating that local abnormal regions with weak boundaries and diverse appearances are difficult to localize. 
Relation- and state-dependent categories such as \textit{Queuing}, \textit{Talking}, and \textit{Waiting} also remain challenging, as they require reasoning about group relations, human states, and surrounding context rather than local object appearance alone.

\textbf{The AP$_{50}$--AP$_{75}$ gap highlights the difficulty of precise region localization.}
Many methods achieve much higher AP$_{50}$ than AP$_{75}$, suggesting that detectors can find approximate region locations but struggle with precise spatial extents. 
For example, DiffusionDet achieves 44.5 AP$_{50}$ but only 22.3 AP$_{75}$, while Co-DETR achieves 46.2 AP$_{50}$ but 26.2 AP$_{75}$. 
Since overall AP averages performance over IoU thresholds up to 0.95, it remains much lower than AP$_{50}$, further reflecting the difficulty of precise region localization.

\subsection{Zero-shot/Open-vocabulary Detection Results}

Table~\ref{tab:zero-shot} reports the zero-shot/open-vocabulary results on RegionDet. 
In sharp contrast to the closed-set setting, all evaluated vision-language detectors struggle severely on Region Detection. 
The best overall AP is only around 1.0, and AP$_{50}$ remains below 3.0 for all models. 
This indicates that current zero-shot/open-vocabulary detectors cannot directly generalize to region targets defined by visual states, relations, scene functions, and contextual cues.

\textbf{This poor performance is consistent across detector families and backbone scales. }
MMGroundingDINO, YOLO-World, YOLOE, and LLMDet all obtain extremely low AP scores, and larger backbones do not consistently bring clear improvements. 
For example, larger variants of MMGroundingDINO and LLMDet show no obvious advantage over their smaller counterparts. 
This suggests that the limitation is not merely caused by model capacity, but is closely related to the object-centric training and evaluation paradigm of existing vision-language detectors.

\textbf{The category-wise results further support this observation. }
Some methods obtain slightly higher scores on \textit{Construction}, possibly because construction scenes contain recognizable object cues such as workers, equipment, or materials. 
However, the overall performance remains very low. 
For relation-defined categories such as \textit{Queuing} and \textit{Talking}, and state- or abnormality-defined categories such as \textit{Damage}, zero-shot/open-vocabulary detectors almost fail to localize valid regions. 
These categories cannot be reliably detected by simply matching text queries to individual object instances, since their spatial extents are determined by group relations, visual states, and scene context.

\textbf{These results reveal a clear gap between open-vocabulary object detection and Region Detection.}
Although existing vision-language detectors have shown strong generalization on object-like categories, they remain strongly biased toward discrete and well-bounded object instances. 
RegionDet therefore provides a challenging benchmark for evaluating whether future open-vocabulary detectors can move beyond object-centric localization toward context- and relation-aware region understanding.

\subsection{Key Findings}

Based on the above results, we summarize three key findings.

\textbf{First, Region Detection is learnable but remains challenging.}
Closed-set detectors can learn region targets under supervised training, showing that Region Detection is a feasible detection task within the standard detection framework. 
However, even recent strong detectors achieve only moderate performance, indicating that region targets pose new challenges beyond conventional object instance detection.

\textbf{Second, region targets require context- and relation-aware understanding.}
The large category-wise performance gap shows that detectors handle visually salient regions better than weak-boundary, state-dependent, or relation-defined targets. 
Categories such as \textit{Damage}, \textit{Waiting}, and \textit{Queuing} require models to reason about visual states, spatial layouts, group relations, and surrounding context, rather than relying only on local object appearance.

\textbf{Third, current open-vocabulary detectors remain strongly object-centric.}
Zero-shot/open-vocabulary detectors perform poorly across all region categories, even with larger backbones. 
This suggests that existing vision-language detectors, although effective for object-like categories, are not yet sufficient for localizing region targets defined by states, relations, functions, and context.




\begin{figure}[t!]
	\centering
	\includegraphics[width=1.0\linewidth]{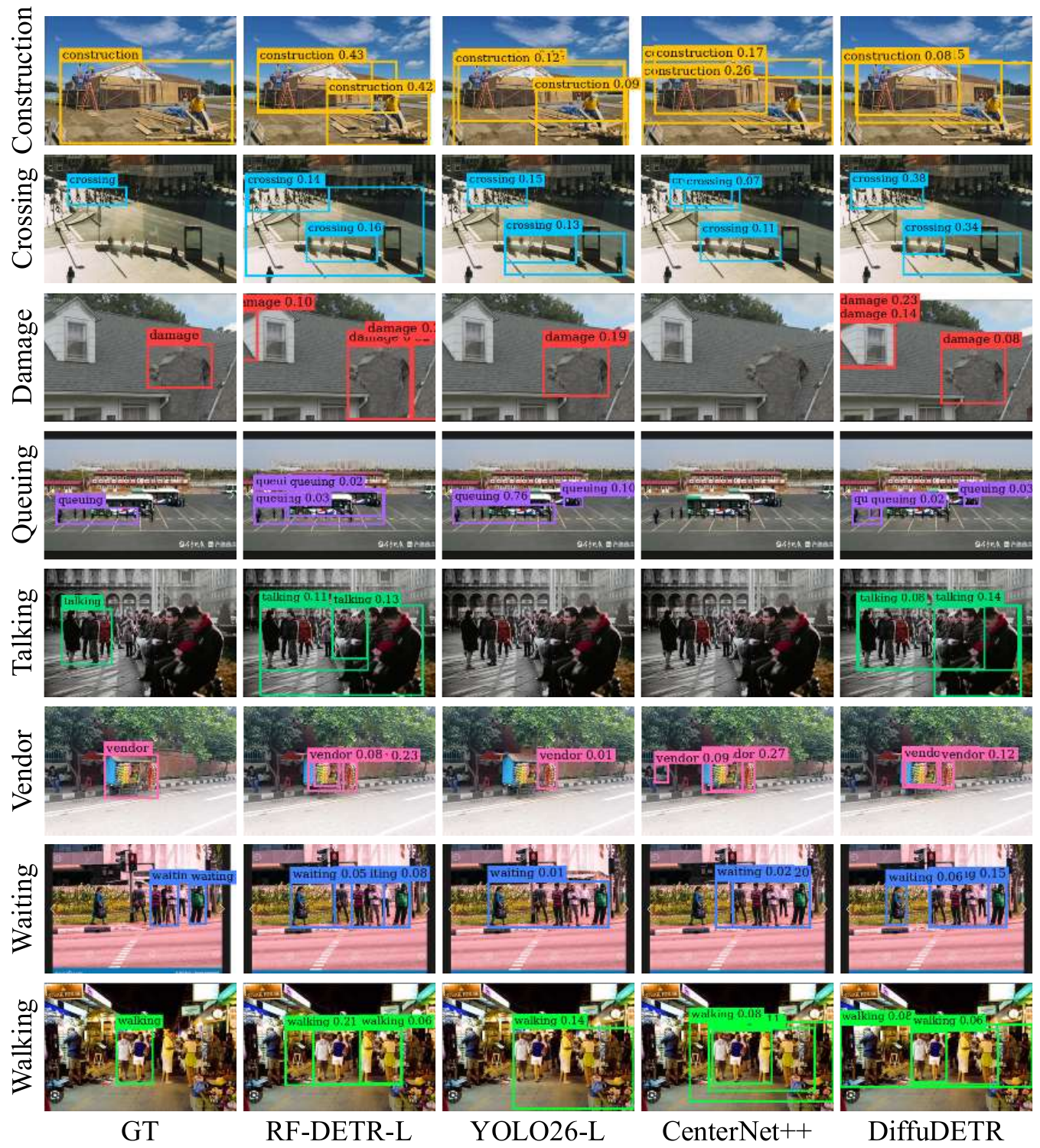}	
    \vspace{-20pt}
    \caption{Qualitative comparison of representative closed-set detectors on RegionDet. 
    The compared methods cover query-based Transformer, dense one-stage, point/center-based, and diffusion-based detectors. 
    While supervised detectors can produce reasonable region predictions, they often show incomplete or imprecise localization.}
    \label{fig:qualitative_closed}
	\vspace{-5pt}
\end{figure}

\begin{figure}[t!]
	\centering
	\includegraphics[width=1.0\linewidth]{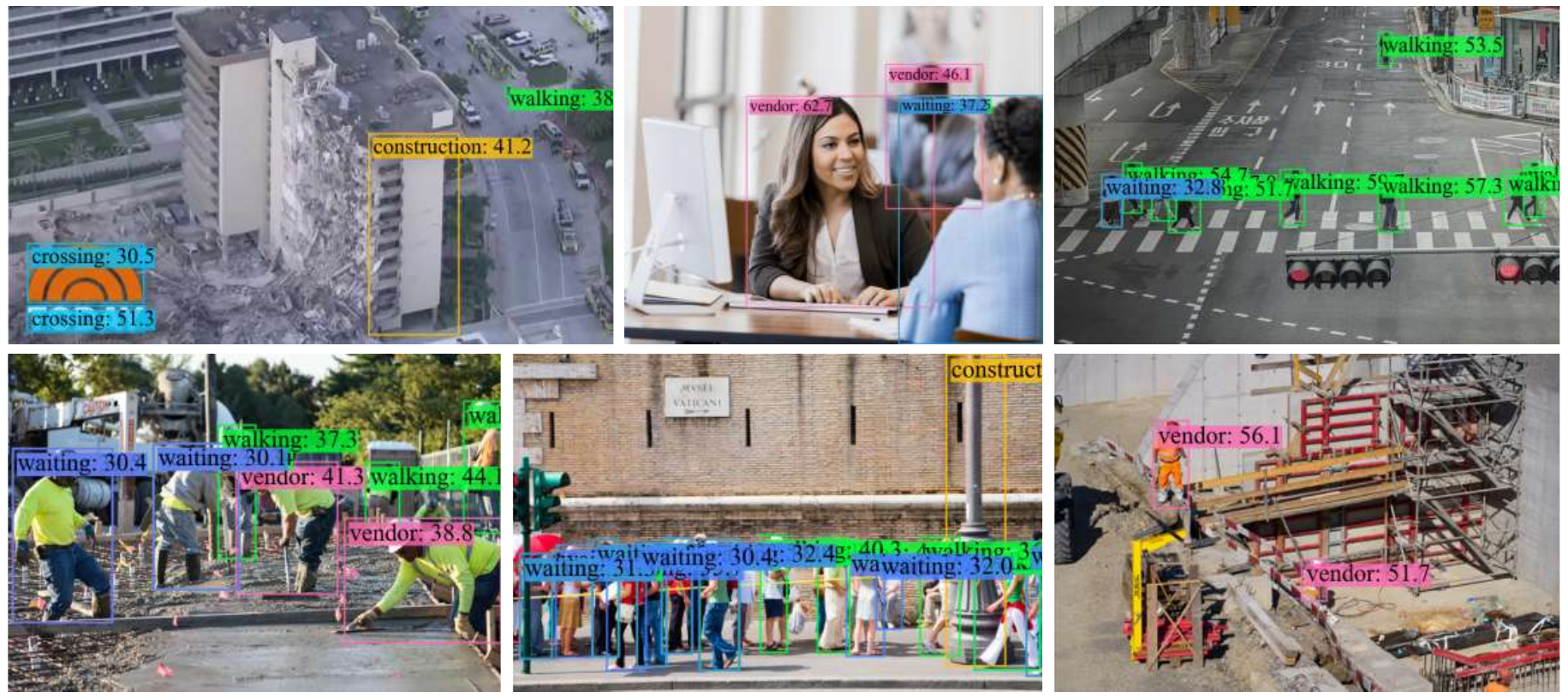}	
    \vspace{-20pt}
    \caption{Qualitative results of YOLOE (v11-L) under the zero-shot/open-vocabulary setting on RegionDet. 
    Although YOLOE (v11-L) can detect many object-like components, its predictions are mostly attached to individual objects or local parts and fail to cover the intended region-level extents.}
    \label{fig:qualitative_zero}
	\vspace{-15pt}
\end{figure}

\subsection{Qualitative Analysis}

\textbf{Closed-set detectors.}
Fig.~\ref{fig:qualitative_closed} shows qualitative comparisons of representative closed-set detectors on RegionDet. 
The compared methods cover query-based Transformer, dense one-stage, point/center-based, and diffusion-based detectors. 
Overall, the visual results are consistent with the quantitative results in Table~\ref{tab:close}. 
With supervised training, detectors can identify relevant region areas to some extent, but their predictions are often partial, shifted, or spatially imprecise.

For context-dependent categories such as \textit{Construction} and \textit{Vendor}, detectors tend to focus on salient components, such as workers, equipment, goods, or pedestrians, but fail to cover the complete functional region. 
For weak-boundary categories such as \textit{Damage}, predicted boxes are often unstable or shifted from the damaged area, reflecting the difficulty of localizing abnormal regions with ambiguous boundaries. 
For relation- and state-defined categories such as \textit{Queuing}, \textit{Talking}, and \textit{Waiting}, detectors sometimes localize individual people or subgroups rather than the full region formed by group relations or human states.

Among the visualized methods, RF-DETR-L generally produces more reasonable region predictions, while other paradigms more frequently show fragmented or partial localization. 
Nevertheless, even stronger detectors still struggle to consistently predict complete region extents. 
These observations suggest that Region Detection requires not only recognizing salient object instances, but also reasoning about weak boundaries, spatial relations, human states, scene functions, and surrounding context.



\textbf{Zero-shot/open-vocabulary detectors.}
Fig.~\ref{fig:qualitative_zero} shows qualitative results of YOLOE (YOLOv11-L), the best-performing zero-shot/open-vocabulary detector in our experiments. 
YOLOE can detect many object-like components, such as people, workers, goods, and local structures, but its predictions are mostly attached to individual objects or local parts rather than complete region targets. 
This is especially evident for activity-, relation-, and context-defined categories such as \textit{Crossing}, \textit{Waiting}, \textit{Walking}, \textit{Construction}, and \textit{Vendor}, where the target extent is determined by human states, spatial layouts, scene functions, and surrounding context. 
These results suggest that current vision-language detectors are not simply limited by object recognition, but by region-level localization and contextual aggregation. 
This qualitative evidence further supports our quantitative results, revealing their strong object-centric bias in localizing regions defined by states, relations, scene functions, and context.

\section{Conclusion and Discussion}
In this paper, we introduce \textbf{Region Detection}, a detection task that extends conventional object detection beyond discrete and well-bounded object instances. 
We construct \textbf{RegionDet}, a benchmark dataset with eight representative region categories, COCO-compatible bounding-box annotations, and standard evaluation protocols. 
RegionDet provides a testbed for localizing visual regions defined by states, relations, scene functions, and contextual cues.
Through systematic experiments, we show that existing detectors face clear challenges on RegionDet. 
Closed-set detectors can partially learn region-level patterns under supervision, but still struggle with weak boundaries, context-dependent extents, and relation-defined targets. 
Zero-shot/open-vocabulary detectors perform poorly, indicating that current vision-language detectors remain strongly object-centric and limited in localizing region targets beyond object instances.

\textbf{Future Directions.}
RegionDet opens promising directions for context- and relation-aware detection, open-vocabulary localization of state-, activity-, and function-defined regions, and more flexible annotation forms for ambiguous region extents. 
We hope RegionDet encourages detection models to move beyond object-centric localization toward richer region-level visual understanding.



\bibliography{aaai2027}


\end{document}